# Persian Semantic Role Labeling Using Transfer Learning and BERT-Based Models

**Saeideh Niksirat Aghdam**

Data Mining Lab, School of Computer Engineering, Iran University of Science & Technology, Tehran, Iran

**Sayyed Ali Hossayni**

Data Mining Lab, School of Computer Engineering, Iran University of Science & Technology, Tehran, Iran

**Erfan Khedersolh Sadeh**

Data Mining Lab, School of Computer Engineering, Iran University of Science & Technology, Tehran, Iran

**Nasim Khozouei**

Department of computer Engineering, Faculty of Engineering, Yasouj University, Yasouj, Iran

**Behrouz Minaei Bidgoli**

Data Mining Lab, School of Computer Engineering, Iran University of Science & Technology, Tehran, Iran

**Correspondence:**
Behrouz Minaei Bidgoli, Department of Computer Engineering Iran University of Science and Technology, Tehran, Iran
**E-mail**: b_minaei@iust.ac.ir

## Abstract

Semantic role labeling (SRL) is the process of detecting the predicate-argument structure of each predicate in a sentence. SRL plays a crucial role as a pre-processing step in many NLP applications such as topic and concept extraction, question answering, summarization, machine translation, sentiment analysis, and text mining. Recently, in many languages, unified SRL dragged lots of attention due to its outstanding performance, which is the result of overcoming the error propagation problem. However, regarding the Persian language, all previous works have focused on traditional methods of SRL leading to a drop in accuracy and imposing expensive feature extraction steps in terms of financial resources, time and energy consumption. In this work, we present an end-to-end SRL method that not only eliminates the need for feature extraction but also outperforms existing methods in facing new samples in practical situations. The proposed method does not employ any auxiliary features and shows more than 16 (83.16) percent improvement in accuracy against previous methods in similar circumstances.

# 1 Introduction

In modern linguistics, natural language processing (NLP) covers different linguistic knowledge and analysis levels, of which "meaning" is a crucial component for both understanding individual words and sentences. Simply put, phonetic or grapheme chains without semantics cannot depict the nature of languages (Palmer *et al.*, 2010). As a result, one of the key NLP topics being discussed is the comprehensive identification of sentence constituents and their association with the subject matter of the sentence, particularly when answering questions such as "who", "what", "why", "how", "when", and "where." This identification process, known as Semantic Role Labelling (SRL), involves assigning semantic roles such as subject, object, and predicate to each word in a sentence (Fei *et al.*, 2021; Jia *et al.*, 2022; Fei *et al.*, 2020).

The detailed study of the level of meaning, or semantics, and SRL plays a crucial role as a pre-processing tool in many NLP applications. These include topic and concept extraction, summarization, machine translation, sentiment analysis, responsive systems, and text mining (Lin *et al.*, 2021). Researchers worldwide generally address the problem of SRL, regardless of the target language and as a shallow semantic parsing problem, using two approaches. In the first approach, referred to as the pipeline method, the problem is divided into two sub-problems: predicate identification and argument role labeling, and each part is addressed independently. This method has two main weaknesses. Firstly, solving each sub-problem independently without considering their connections results in the information obtained in solving the second problem being unable to assist in solving the first problem (Fei *et al.*, 2021). Secondly, any errors that occur in a sub-problem propagate to the subsequent steps, leading to a significant reduction in the efficiency of the algorithm (Lin *et al.*, 2021). Due to these limitations, researchers have widely adopted another technique called end-to-end modeling in recent years (Fu *et al.*, 2019; Shi and Lin, 2019), which often uses deep learning to execute all phases of the SRL process. Although solving the problem using deep learning was first proposed by Colbert et al. in 2011, it did not gain sufficient attention due to its limited competitive results compared to conventional methods (Collobert *et al.,* 2011). However, with the remarkable growth of deep neural networks recently, researchers in the field of NLP have obtained significantly better results in various applications (Radford *et al.*, 2018; Devlin *et al.,* 2019).

SRL has been a highly challenging process, with insignificant improvements (less than 3%) in the F1-score measure of related works over the course of a decade, growing from 77.35% in 2005 (Punyakanok *et al.*, 2005) to 79.4% in 2015 (FitzGerald *et al.*, 2015). This demonstrates the difficulty of the task (He *et al.*, 2017). The main challenges that researchers have faced in solving the problem of SRL can be summarized as follows:

1) For a long time, syntactic analysis was considered one of the pre-processing steps for SRL. But this issue was somewhat challenging because, despite the same syntactic parsing tag for some phrases in different sentences, the same phrases have different semantic roles. In other words, there is no fixed mapping between syntactic analysis tags and semantic role labels, and the meaning of the predicate (verb, noun, or adjective) around which the SRL is done must be understood by the machine to achieve better results. As it is illustrated by

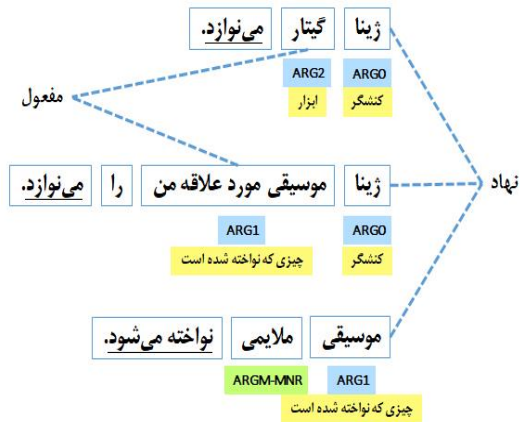

Fig. 1. Lack of fixed mapping between syntactic analysis and semantic role labeling

Fig. 1, in the first and second sentences, the phrases are "گیتار"[1] and "موسیقی مورد علاقه من"[2] tagged as "مفعول"[3] in the syntactic analysis; but in the SRL are labeled as "ابزار"[4] and "چیزی که نواخته شده است"[5]. Also, despite the identification of the expressions "ژینا" and "موسیقی"[6] as "نهاد"[7] in second and third sentences, in syntactic analysis, these two words are labeled as "کنشگر"[8] and "چیزی که نواخته شده است"[9] in semantic role labeling, respectively (He *et al.*, 2017).

Two other issues that pose significant challenges to semantic role labeling are:

2) Determining how to handle prepositions such as "با"[10], and "از"[11]. For example, in the sentences "من با لذت پاستا خوردم" and "من با قارچ پاستا خوردم" which mean "I ate pasta with pleasure" and "I ate pasta with mushrooms", respectively; the phrase after the word "با" in the first sentence describes the state of the verb "خوردن"[12], while the phrase after the word "با" in the second sentence refers to pasta as something that has been eaten (He *et al.*, 2017).

3) The dependency of long phrases within a sentence or even text creates another challenge. As shown in Fig. 2, the analysis of the first sentence is simpler for identifying its semantic roles due to the short distance between the proposition and its arguments. However, in the second sentence, recognizing "ما"[13], as an argument is difficult due to the distance between the proposition "سفر"[14].

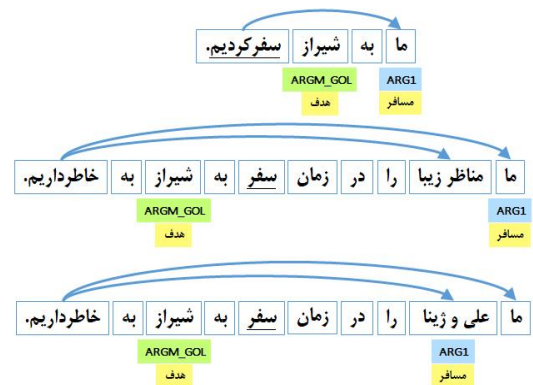

Fig. 2. Long dependencies and the necessity of disambiguation in semantic role labeling

The third sentence presents another type of challenge, as it requires disambiguation. An efficient SRL model should be able to distinguish between words such as "ما", or "we", and "Ali and Jina,", people who are capable of traveling, and a phrase like "مناظر زیبا"[1] (He *et al.*, 2017).

---

1. in English: Guitar
2. in English: my favorite music
3. in English: object
4. in English: tool
5. in English: What's played
6. in English: Music
7. in English: subject
8. in English: actor
9. in English: What's played
10. in English: With
11. in English: Of
12. in English: Eating
13. in English: We
14. in English: Travel

As discussed previously, SRL methods in NLP can be divided into two categories: pipeline methods (traditional) and methods based on deep learning. The proposed method in this research falls under the latter category.

The rest of this paper is organized as follows: Part 2 presents related works while Part 3 explains the proposed method in detail. Lastly, Part 4 covers the evaluation results and provides a comparison of the proposed method with other methods.

## 2 Related Works

### 2.1 Existing corpora in the Persian language for evaluation of algorithms

**2.1.1 Syntactic Valency Lexicon for Persian Verbs:** Among all the corpora and datasets developed in the Persian language, one example is the Syntactic Valency Lexicon for Persian Verbs. This collection contains information related to the syntactic valence of more than 4,500 verbs in Persian. In this lexicon, both mandatory and optional complements of simple, compound, and prefixed verbs are specified. According to the dependency theory, the basic structure of a sentence can be obtained from the valence structure of the verb of that sentence. This highlights the importance of understanding valence constructions of verbs in linguistic texts (Rasooli *et al.*, 2011).

**2.1.2 Persian Syntactic Dependency Treebank:** Another notable corpus is the Persian Syntactic Dependency Treebank, which is the first syntactic corpus of the Persian language. It includes about 30,000 tagged sentences that incorporate syntactic information based on dependency. The dependency features in this syntactic structure have had satisfactory results in machine learning and are well-suited to the nature of disordered languages such as Persian. The sentences were taken from various sources of contemporary Persian texts and all the sentences are labeled with syntactic relationships (based on dependency), such as subject, object, predicate, suffix, and appositive. Additionally, all sentences are tagged with lexical information (parts of speech), including verbs, nouns, adjectives, adverbs, and pronouns. A team of experienced linguists tagged the sentences, which were then reviewed in several stages. The corpus data is provided based on the format of the Computational Linguistics Conference and Natural Language Processing on dependency corpora (Rasooli *et al.*, 2013).

**2.1.3 The Corpus of Semantic Roles of Persian Language:** The Corpus of Semantic Roles of the Persian Language has augmented the semantic layer to the syntactic dependency treebank of Persian language (Rasooli *et al.*, 2013). This dataset includes approximately 30,000 sentences with determined semantic roles for verbs, predicate nouns, and predicate adjectives. Furthermore, the dataset comprises non-subject structures (such as verbal information) along with subject structures (Mirzaei *et al.*, 2015). It's worth noting that, despite its title, this output is a dataset rather than a true corpus, similar to the CoNLL2009 dataset (Hajic *et al.*, 2009).

### 2.2 Existing models for SRL in Persian language

Research regarding SRL in the Persian language has been limited, with most studies relying on traditional methods and using syntactic analysis information to extract features. One example of an algorithm in this field is that of Sadr Mousavi and Shams Fard in 2007 (Sadr Musavi and Shams Fard, 2007), which is a rule-based algorithm using approximately sixty manually defined rules to classify and identify semantic role labels. In 2009, Ghalibaf introduced a memory-based (similarity-based) learning algorithm (Kamal Ghalibaf, 2009) that uses the k-nearest-neighbors algorithm in the learning phase. The data set used in this work has 2000 sentences selected from the corpus of the Research Center for Intelligent Sign Processing for 50 frequent simple verbs in the Persian language, and it has been manually labeled. Another example is the research of Jafarinejad and Shamsfard in 2012 (Jafarinejad and Shamsfard, 2012), which relies on rule-based methods to prepare a system for recognizing semantic roles through the morphology of words. In the same year, Saeedi and Faili (Saeedi and Faili, 2012) used a learning algorithm based on memory and k-nearest neighbors and used 1000 sentences from Bijan Khan corpus containing 140 verbs as training data. Their goal in that research was to find a set of features that would get the best result. According to the research, these features are the root of syntactic analysis, the tag of parts of speech, the label of the named entity of the root of syntactic analysis, the root of the verb, and the syntactic capacity of the verb. In 2016, Rezai Sharifabadi and colleagues (Rezaei Sharif Abadi and Khosravi Zade Forooshani, 2016) carried out complete syntactic analysis based on dependency and identified the predicate (verb) of each sentence. They then created a feature vector for each word in the sentence, consisting of the root of the verb, the direction of the verb, the tag of part of speech, the path of the syntactic parsing tree, and other factors. Then, the Naive Bayes and maximum entropy algorithms to train the tagger. In a more recent work, Lazemi and colleagues (Lazemi et al., 2019) treat SRL as a classification problem and use complete dependency analysis for each proposition in the sentence. They then extract feature vectors for each argument and use the maximum entropy ranking to determine the labels of the arguments. The same researcher published another paper in 2020 (Lazemi et al, 2020) which employs deep neural networks. The aim of the paper is to use minimal information of the dependency decomposition tree. In this regard, first, words of a sentence are mapped to the vector space using the Word2Vec (Mikolov et al., 2013) algorithm, then the syntactic features of each word are extracted, and in the next step, using a Long-short-term-memory (LSTM) neural network layer, the syntactic dependency features of each proposition-argument pair, which is the dependency decomposition tree path, is extracted. Finally, all extracted features are combined and given as input to a bi-directional Long-short-term memory (BiLSTM) neural network. At the end, the arguments are classified using the Softmax classifier. In the researches (Lazemi et al., 2019; Lazemi et al, 2020), two data sets of Persian language semantic roles (Mirzaei et al., 2015) and a smaller data set (Lazemi et al, 2018), which is produced by the same researchers by adding a layer of semantic roles on the Uppsala data set (Seraji, et al., 2016), were utilized for training and testing purposes. Another recent research is the research of Mirzaei and his colleagues in 2020 (Mirzaei et al., 2020). In this research, due to the use of semantic roles of the Persian language (Mirzaei et al., 2015) as training data, and as a result, the specificity of propositions, only the identification of arguments and their classification is done. In this algorithm, after extracting the required features from the data set, the arguments are identified using the maximum entropy, and the support vector machine algorithm is also used in the classification section. In 2021, a paper was published by Shojaei and colleagues, which the corresponding method was employing FastText (Bojanowski *et al*., 2017), for the embedding step, and the utilized tool as classifier was BiLSTM neural network (Shojaei Baghini *et al*., 2021).

---
[1]. In English: Beautiful Scenery

It can be easily understood that the majority of works employed traditional methods and deep learning has only been used in two papers [25,31].

In traditional and rule-based methods, there is a need for features extraction, which leads to a lot of pre-processings before the main process, if these pre-processings are done by machine, their errors will propagate in the main process and reduce the accuracy. Also, obtaining these auxiliary features in a manually and golden form leads to significant financial costs.

On the other hand, the progress of machine learning algorithms and the use of deep neural networks have led to the elimination of many pre-processings and as a result, reducing the dependence on manual extraction of features, which caused significant improvements in the results of the algorithms since 2015 (Zhou and Xu, 2015). Although two studies (Lazemi *et al*, 2020; Shojaei Baghini *et al*., 2021) utilized deep learning for classification, word embedding in these works was done using Wor2Vec and FastText algorithms, respectively. The drawback of these embedding algorithms is that they are static; In other words, the vectors to which the words are mapped are always the same and have nothing to do with the context of the text in which the word appeared. For example, consider the word "خودرو" in two sentences "سرخس گیاهی[1]" and "[2] صنعت خودرو با مشکل خودرو است. مواجه شده است.". Although the writing of the words is identical, meanings are considerably far. In the first sentence, the word means "self-grow" or "wild" and in the second sentence it means "vehicle". Due to neglecting contextual information, the vector created by the above methods for both words are the same. In another example, consider two words "Men" and "Women", there can cases that these words are used to mention contrasts, whilst there can be other cases that they are used to mention human species and the similarities. Since 2017 and in order to solve this problem, methods of "contextual word embedding" were proposed.

# 3 Structure of the Proposed Algorithm

One of the most effective methods for contextual word embedding in NLP is the BERT [11] embedding. In this study, we have designed an algorithm to solve the problem of SRL, which is based solely on the BERT contextual embedding and "fine tuning", without using other auxiliary features. Our algorithm employs deep learning techniques and follows the Propbank style, using an adapted version of BERT contextual embedding for the Persian language such as MultiLingualBERT (Devlin *et al.*, 2019) and ParsBERT (Farahani *et al*., 2021). The algorithm processes raw text without

---

[1]. The Fern is a wild plant.

[2]. The automotive industry is in trouble.

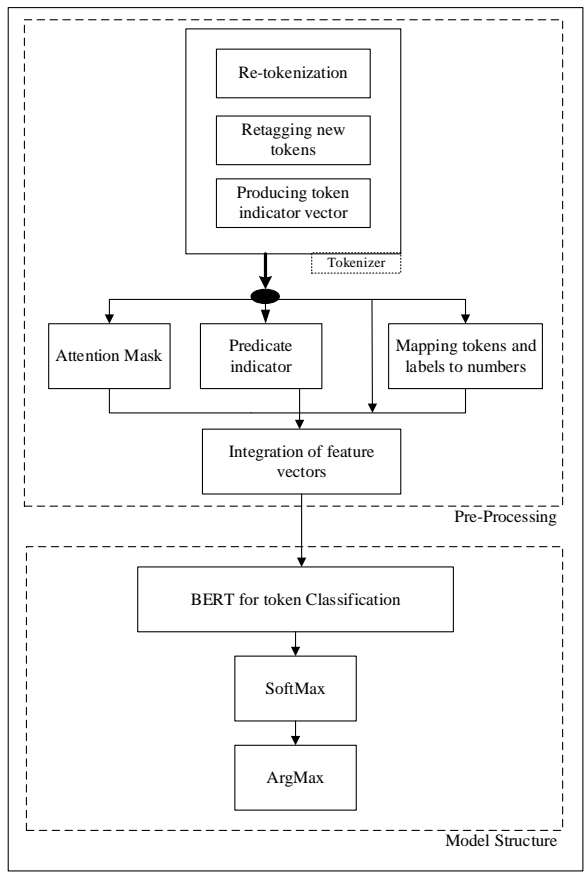

Fig. 3. General stages of the proposed method

additional features and performs SRL in a dependency-based manner. The generalized steps of our algorithm are shown in Fig. 3 and are explained in detail in the following:

**3.1 Pre-processing**

The proposed algorithm requires four types of information in the form of lists containing numerical vectors to perform SRL operations:

1. A list containing a tokenized sentence using the BERTTokenizer library; plus beginning and ending tokens.

2. A list that includes labels for tokens.
3. A list that indicates the position of the predicate.
4. A list that is used as attention mask.

The following section will describe how these vectors are formed.

**Data Tokenization**

To prepare the sentences for BERT-based SRL, we utilize the BERTTokenizer library to tokenize and convert the sentences into vectors. The BERTTokenizer library applies the WordPiece algorithm to tokenize the sentences into tokens. By referencing a file called Vocab.txt, which is available in the HuggingFace repository, the library identifies tokens. If a word does not exist in the file, it is converted into sub-words. This process continues until every word or its sub-words are identified in the vocabulary file. If a word or its sub-words do not exist in the vocabulary file, it is considered unknown. After re-tokenization process, in order to indicate beginning and end of each sentence, reserved tokens [CLS] and [SEP] are added at the beginning and end of the sentence, respectively.

During this tokenization process, the primary tags assigned to each token are revised based on the separation of main words and sub-words. Thus, the main words are tagged by their previous tags, while the sub-words are labeled as "X". Table 1 provides an example of this tokenization process for the sentence:

"در ابتدای این مانور آزمایشـی نشـسـت و برخاسـت هلی‌کوپتر و عملیـات راپل (پیـاده کردن تکاورهـای ویژه) اجرا شد."

Which means "At the beginning of this trial maneuver, the helicopter took off and landed, and the rappel operation (launching of special rangers) was carried out."

It is noteworthy that this example focuses on the labeling of the predicate "شدن".

Table 1. An example of sample retokenization

| The prepared sample with token labels | | Token indicator | Re-tokenized sample alongside new labels | |
|---|---|---|---|---|
| در | B-TMP | 1 | در | B-TMP |
| ابتدای | O | 1 | ابتدای | O |
| این | O | 1 | این | O |
| مانور | O | 1 | مانور | O |
| آزمایشی | O | 1 | [UNK] | O |
| نشست | B-A1 | 1 | نشست | B-A1 |
| و | O | 1 | و | O |
| برخاست | O | 1 | برخاست | O |
| هلی‌کوپتر | O | 1 | [UNK] | O |
| و | O | 1 | و | O |
| عملیات | O | 1 | عملیات | O |
| راپل | O | 1 | راپ | O |
|  |  | 0 | ##ل | X |
| ) | O | 1 | ) | O |
| پیاده | B-PRD | 1 | پیاده | B-PRD |
| کردن | O | 1 | کردن | O |
| تکاورهای | O | 1 | تکاور | O |
|  |  | 0 | ##های | X |
| ویژه | O | 1 | ویژه | O |
| ( | O | 1 | ( | O |
| اجرا | B-NVE | 1 | اجرا | B-NVE |
| شد | B-V | 1 | شد | B-V |
| . | O | 1 | . | O |

Table 2. Labels mapped in the case that the data set is given to the algorithm only with verb statements

| Label | Number | Label | Number |
|---|---|---|---|
| [UNK] | 0 | [PAD] | 1 |
| B-TMP | 2 | O | 3 |
| X | 4 | B-A1 | 5 |
| B-NVE | 6 | B-PRD | 7 |
| B-A0 | 8 | B-V | 9 |
| B-CAU | 10 | B-A2 | 11 |
| B-DIS | 12 | B-MNR | 13 |
| B-EXT | 14 | B-ADV | 15 |
| B-PRP | 16 | B-LOC | 17 |
| B-INS | 18 | B-A4 | 19 |
| B-CON | 20 | B-A3 | 21 |
| B-COM | 22 | B-RPT | 23 |
| B-DIR | 24 | B-GOL | 25 |
| B-ASP | 26 | B-MOD | 27 |
| B-RCL1 | 28 | B-Def | 29 |
| B-REC | 30 | B-Top | 31 |
| B-NEG | 32 | B-ArgA | 33 |
| B-ADJ | 34 | B-NEG | 35 |
| B-T1 | 36 | B-Arg2_B0 | 37 |
| B-Arg2_B1 | 38 | B-PUNC | 39 |

## Mapping Semantic Role Tags and Morphemes to Numbers

The defined semantic role labels, which are classes in the dataset, and the reserved tokens are extracted and then mapped to respective numbers. These numbers are then used to convert lists that contain semantic role labels into vectors. Table 2 provides an example of this mapping process when the data set only has verb predicates.

## Predicate Indicator Vector

To ascertain the position of the predicate in a given sentence, a list with the same length as the list of tokens is generated. In the list, the position of the primary predicate is assigned a value of 1, while remaining predicates are assigned a value of 0. This approach enables the unambiguous identification of primary predicate of the sentence.

## Attention Vector

To enable the algorithm to recognize the boundaries of tokens within a sentence, we create a list of the lengths of tokens in the sentence. Each entry in this list is assigned a value of 1, and the list is then padded with zeroes from the right side to ensure a uniform length across all sentences in the dataset, according to the length of the longest sentence. This standardized format allows the sentences to be processed by the neural network.

### 3.2 Structure of Neural Network

In this section, a pre-trained BERT model suitable for the Persian language is selected, as word embedding algorithm, from the Models section on Huggingface.co, with options including HooshvareLab/bert-fa-base-uncased and bert-base-multilingual-cased. The default choice for our algorithm is the former, although other similar models can also be implemented. Once the word embedding model is selected, we perform the fine-tuning operation on it by injecting the feature vectors prepared in the previous phase.

As SRL involves a multiple classification problem for the classification of the tokens within the sentence, the most suitable BERT model is the BERTForTokenClassification model with added token classification layers.

The predictions resulting from BERTForTokenClassification are normalized using the softmax function, which produces a set of numbers for each token that are assigned to the classification classes, with the sum of these normalized vectors being 1. The normalized vectors have a range between [0,1], as opposed to the initial interval of [-∞, ∞]. Finally, the argmax function selects the class with the highest number as the final result of the model.

# 4 Evaluation of the Proposed Method and Analysis of its Performance

### 4.1 Dataset Preprocessing

The employed dataset in this work is the Persian Language Semantic Roles Corpus (Mirzaei *et al*., 2015), which has undergone cleaning prior to its use. This corpus was created by applying a semantic layer to the syntactic dependency treebank of Persian language developed by Rasouli and his colleagues (Rasooli *et al*., 2013), and contains approximately 30,000 tagged sentences. In this corpus, all semantic roles related to verbs, predicate nouns, and predicate adjectives within each sentence were identified.

The Persian Language Semantic Roles Corpus is similar to CoNLL2009 (Hajic *et al*., 2009). As shown in Table 3, from left to right, the columns in this dataset correspond to token counter, Form (token), gold standard lemma, automatically predicted lemma, gold standard POS, automatically predicted POS, gold standard morphological features, ID of gold standard syntactic head, automatically predicted syntactic head, column 9 contains "Y" for argument-bearing tokens, identifier of a semantic "predicate", and from column number 11 to the last column, each column is dedicated to the semantic role label of a proposition in the sentence.

The dataset-cleaning process was done in two steps, which involved correcting any noise and removing outliers. The following rules were applied during this operation:
• Ensuring the data structure complies with the structure outlined in Table 3.
• Ensuring that the correct number of semantic role label columns exist for each identified proposition.

Table 3. An example of an annotated sentence in the body of semantic roles of the Persian language

| ID | FORM | LEMMA | POS | PPOS | FEAT | HEAD | PHEAD | FILLPRED | PRED | APRED1 | APRED2 |
|---|---|---|---|---|---|---|---|---|---|---|---|
| 1 | حاضران | حاضران | N | ANM | | 12 | SBJ | _ | _ | _ | **A0** |
| 2 | با | با | PREP | PREP | | 12 | ADV | _ | _ | _ | **MNR** |
| 3 | اجرای | اجرا | N | IANM | | 2 | POSDEP | Y | اجرا | _ | _ |
| 4 | سرود | سرود | N | IANM | | 3 | MOZ | _ | _ | A1 | _ |
| 5 | ای | ای | ADR | PRADR | | 4 | MOZ | _ | _ | _ | _ |
| 6 | ایران | ایران | N | IANM | | 5 | POSDEP | _ | _ | _ | _ |
| 7 | محمد | محمد | N | ANM | | 5 | MOZ | _ | _ | _ | _ |
| 8 | اصفهانی | اصفهانی | N | ANM | | 7 | MOZ | _ | _ | _ | _ |
| 9 | از | از | PREP | PREP | | 12 | VPP | _ | _ | _ | **A1** |
| 10 | جای | جای | N | IANM | | 9 | POSDEP | _ | _ | _ | _ |
| 11 | خود | خود | PR | CREFX | | 10 | MOZ | _ | _ | _ | _ |
| 12 | برخاستند | بر#خاست#خیز | V | ACT | | 0 | ROOT | Y | بر#خاست#خیز | _ | _ |
| 13 | . | . | PUNC | PUNC | | 12 | PUNC | _ | _ | _ | _ |

• Inserting an underscore in cells that were empty.
• Replacing the ".." with an underscore that appeared in some cells by mistake.
• Replacing the empty spaces in monograms and lemmas with an underscore.
• Ensuring that there is at least one proposition per sentence.
• Removing sentences without any semantic role annotations.
• Removing distorted, destroyed, and unreadable sentences.
• Removing sentences with unreasonable propositions.
The results of the cleaning process are presented in Table 4.

Table 4. The results of cleaning the dataset

| | |
|---|---|
| Number of all sentences in the dataset | 30005 |
| Number of reviewed sentences | 84 |
| Number of deleted sentences | 24 |
| Number of sentences used as input data | 29981 |

### 4.2 Preparing the Required Data for Model Training and Evaluation

At this stage, the dataset is prepared. If more than one predicate is identified within a sentence and more than one semantic role label column exists for that sentence, a separate sample is considered in the dataset for each predicate. An example of this process is shown in Table 5 for the sentence described in Table 3.

After performing the initial processing of the data, a sample is created for each predicate that includes three main features. Firstly, the "seq_words" property is generated, which is a list of the tokens that comprise the sentence. Secondly, the "BIO" property is generated, which is a list of semantic role tags. Finally, the "pred_sense" property is generated, which is a list of sub-words that represent a sense of the predicate and its location in the sentence.

During the pre-processing stage, it is important to determine whether the resultant samples should contain only verb statements or if they should include all identified propositions, including verbs, nouns, and adjectives. Table 6 provides the number of resulting samples from this step, including samples that contain only verbs and those that include all identified propositions.

Table 5. Generating separate samples for different predicates in a sentence

| Sample | Feature name | Feature value |
|---|---|---|
| 1 | Seq_Words | ["حاضران", "با", "اجرای", "سرود", "ای", "ایران", "محمد", "اصفهانی", "از", "جای", "خود", "برخاستند", "."] |
| | BIO | ["o", "o", "B-N-V", "B-A1", "o", "o", "o", "o", "o", "o", "o", "o", "o"] |
| | Pred_Sense | ["اجرا", 2] |
| 2 | Seq_Words | ["حاضران", "با", "اجرای", "سرود", "ای", "ایران", "محمد", "اصفهانی", "از", "جای", "خود", "برخاستند", "."] |
| | BIO | ["B-A0", "B-MNR", "o", "o", "o", "o", "o", "o", "B-A1", "o", "o", "o", "B-V", "o"] |
| | Pred_Sense | ["خیز", 11] |

To prepare the data for the training, validation, and test processes, the sklearn library was utilized and the K-Folds cross-validation method with the parameter k=10 was employed. In this process, the samples were divided into three sets, where 80% of the samples were assigned to the training data, 10% to the validation data, and the remaining 10% to the test data. The trained algorithm was then tested on different

Table 6. The results of data preparation

| | |
|---|---|
| Number of sentences used as input data | 29981 |
| Number of samples when only verb predicates are considered | 59917 |
| Number of samples when all predicates are considered | 91804 |

permutations resulting from this operation, with the results being averaged on the F1 score.

### 4.3 Setting the Parameters of the Main Model and the Simulation Platform

The number of input neurons to the network varies depending on the selected word embedding algorithm and the pre-processing steps carried out on the sentences corresponding to that algorithm. For example, if the ParsBERT algorithm is selected, the maximum length of the sentence sequence allowed is 197. Whereas, if the Multilingual-BERT algorithm is used, the maximum length will be 277. The difference in the maximum sentence length can be attributed to the use of a more appropriate dictionary for the Persian language in the ParsBERT.

Network training was conducted using an NVIDIA GeForce GTX 1070 graphics processor with a fixed learning rate. The network optimization was performed using the AdamW algorithm (Loshchilov and Hutter, 2017). The hyperparameters for network training are prov ided in Table 7.

Table 7. Hyperparameters of network training

| Hyperparameter | Value |
|---|---|
| Number of encoder layers | 12 |
| Number of all parameters | 110e-06 |
| Learning rate | 2e-05 |
| Batch size | 16 |
| epoch | 10 |

### 4.4 Results

The proposed model was trained, and evaluated using both MultiLingualBERT and ParsBERT word embedding algorithms in two separated scenarios. Table 8 provides a detailed comparison between the proposed and existing methods in terms of syntax awareness (SA[1]), the number of features employed for training, the number of semantic roles, the method used for word embedding (if needed), Methodology, and F1-score.

Table 8. Comparison of proposed and existing methods

| Algorithm | SA | Number of | | | Word embedding method | Data splitting method | Methodology and steps | $F_1$ |
|---|---|---|---|---|---|---|---|---|
| | | Sentences | Roles | Features | | | | |
| Lazemi *et al.*, 2019 | yes | 29982 | 8 | 8 | Word2vec For clustering | 10-fold cross validation | Argument detection And classification, Maximum entropy, k-means predicate clustering | 74/87 |
| Mirzaei *et al.*, 2020 | yes | 29982 | 33 | 15 to 30 | None (not needed) | 10-fold cross validation | Argument detection, Maximum entropy, SVM Argument classification | 84 |
| Lazemi *et al*, 2020 | yes | 29982 | 8 | None | Word2vec | 10-fold cross validation | Argument detection And classification two-layer BiLSTM Softmax | 68/75 |
| | | | | Word root and part of speech | | | | 72/61 |
| | | | | Syntactic information | | | | 80/01 |
| Shojaei Baghini *et al.*, 2021 | No | 29982 | 33 | 3 | FastText | 5-fold cross validation | Argument detection And classification, BiLSTM ReLU | 88/59 |
| Proposed | No | 29982 | 33 | None | MultiLingual BERT | 10-fold cross validation | Disambiguation of predicates, Argument detection, argument classification, Fine-Tuning, Softmax | 85/19 |
| | | | | | ParsBERT (Base) | | | 86/13 |

---

1. Syntax Aware (SA): employment of syntactic-decomposition information; this concept is in the contrast to Syntax Agnostic which means neglecting that information.

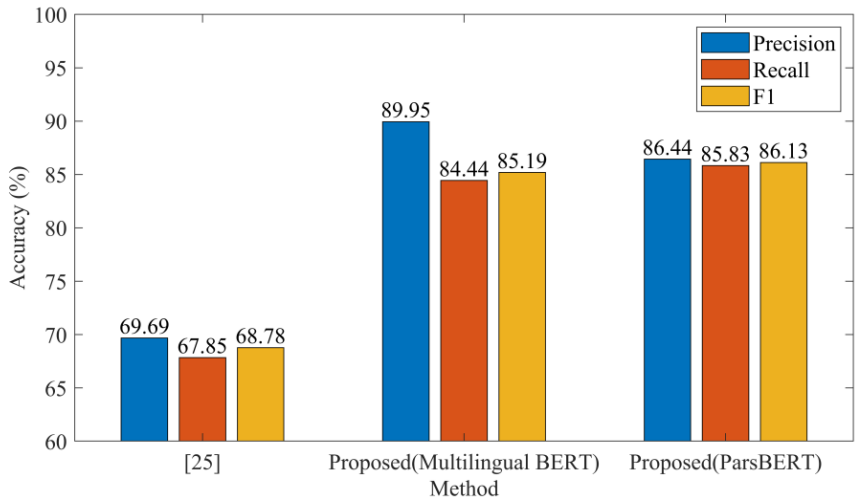

Fig. 4. Comparison of the proposed method to the only existing method in the Persian language that utilizes no auxiliary features

As shown in Table 8, there are other methods with high accuracy in the Persian language, but they all are benefited by features. However, our proposed method achieved significant results with only monogrammed text and final labels, indicating its effectiveness in addressing the SRL problem in the Persian language. Additionally, by comparing the results of the proposed algorithm and (Lazemi et al, 2020), which is the only existing method that does not benefits auxiliary features, the superiority (over 16 percent higher accuracy) of the proposed method is observed. A more detailed comparison is presented by Fig. 4. As can be seen in the results presented in Table 8, the number of classes in algorithms (Lazemi et al., 2019; Lazemi et al, 2020) are smaller, in order to make the results comparable, the reported results were presented as an average over all the predictions made for the classification classes.

To ensure results reliability and avoid the dependence on data-dividing, the method was validated using K-fold cross-validation with the parameter k=10. Using this approach ten different subsets were acquired. each of these subsets was available for model testing, while one subset out of nine remaining subsets (chosen randomly) was dedicated to validation and the rest of subsets (eight of them) were used for model training.

The proposed algorithm was trained, applying subset 1, and tested in two modes, considering only verb propositions or all propositions. Additionally, changing the word embedding algorithm was also considered and both the MultiLingualBERT and ParsBERT algorithms were employed for this purpose in different scenarios. The findings are shown in Table 9. As anticipated, training the algorithm with the ParsBERT, which is specific to the Persian language and was presented by the Hushwarah group (Farahani et al., 2021), produced better results with a difference of around 1% compared to training with the multilingual BERT model.

Furthermore, the proposed algorithm was trained, utilizing the ParsBERT word embedding algorithm and nine other subsets.

Table 9. Performance of the method developed using two different word embedding algorithms (ParsBERT and MultiLingualBERT)

| Utilized Predicates | Word embedding algorithm | Number of | | | Precision | Recall | F1 |
|---|---|---|---|---|---|---|---|
| | | Samples | Sentences | Roles | | | |
| Only verb Predicates | ParsBERT | 59917 | 29981 | 33 | 85/03 | 84/9 | 84/96 |
| All Predicates | ParsBERT | 91804 | 29981 | 33 | 86/26 | 86/27 | 86/26 |
| Only Verb Predicates | MultiLingualBERT | 59917 | 29981 | 33 | 84/75 | 83/27 | 84 |
| All Predicates | MultiLingualBERT | 91804 | 29981 | 33 | 84/95 | 84/44 | 85/19 |

The outcomes of the proposed model for all 10 training subsets are presented in Table 10. As illustrated in the table, the variances of the F1-score were 0.041 and 0.034 for two, "verbs only" and "all predicates" cases, respectively. These results indicate the reliability of performance of the algorithm.

**5 Summary**

In this study, an algorithm for SRL is proposed that does not rely on auxiliary features, such as syntactic analysis information or part-of-speech tags. The algorithm is trained solely using raw text and semantic role labels.

In this approach a BERT-based contextual embedding method is employed for converting text to numerical vector. By using WordPiece library and re-tokenization of the input text, the algorithm can break down unknown tokens to the level of letters, which leads to better performance in operational and real environments. Compared to other studies that use static algorithms for converting text to numerical

Table 10. performance of the proposed method (with ParsBERT word embedding) on all permutations of data

| Utilized Predicates | All Predicates | | | Only Verbs | | |
|---|---|---|---|---|---|---|
| Permutation | Recall | Precision | $F_1$ | Recall | Precision | $F_1$ |
| first | 86/27 | 86/26 | 86/26 | 84/9 | 85/03 | 84/96 |
| second | 85/48 | 86/57 | 86/02 | 84/56 | 86/13 | 85/36 |
| third | 85/66 | 86/48 | 86/05 | 85/13 | 85/08 | 85/01 |
| fourth | 85/81 | 86/66 | 86/23 | 84/25 | 85/55 | 84/9 |
| fifth | 85/81 | 86/87 | 86/36 | 84/45 | 85/04 | 85/04 |
| sixth | 86/29 | 86/5 | 86/39 | 84/82 | 85/47 | 85/15 |
| seventh | 86/17 | 86/25 | 86/21 | 85/37 | 85/05 | 85/21 |
| eighth | 85/64 | 86/1 | 85/87 | 84/37 | 85/08 | 84/72 |
| ninth | 85/6 | 86/12 | 85/86 | 84/48 | 85/55 | 85/01 |
| tenth | 85/6 | 86/59 | 86/09 | 84/88 | 84/52 | 84/7 |
| **Average** | **85/83** | **86/44** | **86/13** | **84/72** | **85/25** | **85/01** |
| **Variance** | **0/91** | **0/063** | **0/34** | **0/129** | **0/192** | **0/041** |

vector, our approach is more flexible and can handle previously unseen words and tokens.

Overall, our proposed algorithm does not rely on complex features or require extensive pre-processing steps. The use of BERT-based contextual embedding and careful handling of unknown tokens make it a promising solution for SRL in NLP.